\documentclass[12pt]{article}

\usepackage{amsfonts}
\usepackage{amsmath}
\usepackage{amssymb}
\usepackage[backend=biber,style=nature]{biblatex}
\usepackage{booktabs}
\usepackage[
    font=sf,
    labelfont=bf,
    figurename=Fig.,
    labelsep=period
]{caption}
\usepackage{graphicx}
\usepackage[detect-all]{siunitx}
\usepackage{times}
\usepackage{tikz}
\usepackage{url}

\newcommand{\Mod}[1]{\ (\mathrm{mod}\ #1)}

\newenvironment{sciabstract}{%
\begin{quote} \bf}
{\end{quote}}

\title{FastMapSVM: Classifying Complex Objects Using the FastMap Algorithm and Support-Vector Machines}

\author{
Malcolm C.~A.~White,$^{1\ast}$
Kushal Sharma,$^{2}$
Ang Li,$^{2}$
T.~K.~Satish Kumar,$^{2}$\\
Nori Nakata$^{1,3}$\\
\\
\normalsize{$^{1}$Massachusetts Institute of Technology,}\\
\normalsize{Cambridge, MA 02139, USA}\\
\normalsize{$^{2}$University of Southern California}\\
\normalsize{Los Angeles, CA 90007, USA}\\
\normalsize{$^{3}$Lawrence Berkeley National Laboratory}\\
\normalsize{Berkeley, CA 94720, USA}\\
\\
\normalsize{$^\ast$To whom correspondence should be addressed; E-mail: malcolmw@mit.edu.}
}

\date{}

\begin{document}

\maketitle

\begin{sciabstract}

Neural Networks and related Deep Learning methods are currently at the leading edge of technologies used for classifying objects. However, they generally demand large amounts of time and data for model training; and their learned models can sometimes be difficult to interpret. In this paper, we advance FastMapSVM\textemdash an interpretable Machine Learning framework for classifying complex objects\textemdash as an advantageous alternative to Neural Networks for general classification tasks. FastMapSVM extends the applicability of Support-Vector Machines (SVMs) to domains with complex objects by combining the complementary strengths of FastMap and SVMs. FastMap is an efficient linear-time algorithm that maps complex objects to points in a Euclidean space while preserving pairwise domain-specific distances between them. We demonstrate the efficiency and effectiveness of FastMapSVM in the context of classifying seismograms. We show that its performance, in terms of precision, recall, and accuracy, is comparable to that of other state-of-the-art methods. However, compared to other methods, FastMapSVM uses significantly smaller amounts of time and data for model training. It also provides a perspicuous visualization of the objects and the classification boundaries between them. We expect FastMapSVM to be viable for classification tasks in many other real-world domains.

\end{sciabstract}

\section*{Introduction}
\label{sec:introduction}

\par
Various Machine Learning (ML) and Deep Learning (DL) methods, such as Neural Networks (NNs), are popularly used for classifying objects. For example, a Convolutional NN (CNN) is used for classifying Sunyaev-Zel'dovich galaxy clusters~\cite{lhcawnt19}, a densely connected CNN is used for classifying images~\cite{hlvw17}, and a deep NN is used for differentiating the chest X-rays of Covid-19 patients from other cases~\cite{entfhmwpn20}. However, they generally demand large amounts of time and data for model training; and their learned models can sometimes be difficult to interpret.

\par
In this paper, we advance FastMapSVM~\cite{bka09}\textemdash an interpretable ML framework for classifying complex objects\textemdash as an advantageous alternative to NNs for general classification tasks. While most ML algorithms learn to identify characteristic features of~\emph{individual objects} in a class, FastMapSVM leverages a domain-specific distance function on~\emph{pairs of objects}. It does this by combining the strengths of FastMap and Support-Vector Machines (SVMs). In its first stage, FastMapSVM invokes FastMap, an efficient linear-time algorithm that maps complex objects to points in a Euclidean space, while preserving pairwise distances between them. In its second stage, it invokes SVMs and kernel methods for learning to classify the points in this Euclidean space. The FastMapSVM framework that we implement in this paper is conceptually identical to the SupFM-SVM method of~\textcite{bka09}. However, the work of~\textcite{bka09} only emphasized the sparse kernel representations enabled by FastMap, whereas our work manifests several additional benefits FastMapSVM offers, even when SVMs themselves are rendered ineffective due to the overwhelming complexity of the objects. From this perspective, our development of FastMapSVM is novel and advantageous in domains with complex objects.

\par
First, there are many real-world domains in which feature selection for~\emph{individual objects} is challenging, but a distance function on~\emph{pairs of objects} is well defined and easy to compute. In such domains, FastMapSVM is more easily applicable than other ML algorithms that focus on the features of individual objects. Examples of such real-world objects include audio signals, seismograms, DNA sequences, electrocardiograms, and magnetic-resonance images. While these objects are complex and may have many subtle features that are hard to recognize, there exists a well-defined distance function on pairs of objects that is easy to compute. For instance, individual DNA sequences have many complex and subtle features, but the~\textit{edit distance}\footnote{The edit distance between two strings is the minimum number of insertions, deletions, or substitutions that are needed to transform one to the other.} between two DNA sequences is well defined and easy to compute.

\par
Second, because FastMapSVM generates a Euclidean embedding, it provides a perspicuous visualization of the objects and the classification boundaries between them. This aids human interpretation of the data and results. It also enables a human-in-the-loop framework for refining the processes of learning and decision making. Moreover, FastMapSVM is able to produce the visualization very efficiently because it invests only linear time in generating the Euclidean embedding.

\par
Third, FastMapSVM uses significantly smaller amounts of time and data for model training compared to other ML algorithms. This is because, given $N$ objects and their classification labels (training instances), FastMapSVM leverages $O(N^2)$ pieces of information via a distance function that is defined on every pair of objects. In contrast, ML algorithms that focus on individual objects leverage only $O(N)$ pieces of information. Despite considering $O(N^2)$ pieces of information to generate a Euclidean embedding, FastMapSVM invests only $O(N)$ time to do so.

\par
Fourth, FastMapSVM extends the applicability of SVMs and kernel methods to domains with complex objects. SVMs and associated kernel methods~\cite{s03} constitute a very powerful ML framework for classification tasks. However, they are generally applicable only when the objects can be represented in a geometric space. As mentioned before, in many real-world domains, it is unwieldy to represent all the features of a complex object in a geometric space. In such domains, FastMapSVM satisfies this geometric requirement by generating a low-dimensional Euclidean embedding of complex objects via a distance function.

\par
In this paper, we demonstrate the efficiency and effectiveness of FastMapSVM in the context of classifying seismograms. In fact, this is a particularly illustrative domain because seismograms are complex objects with subtle features indicating diverse energy sources such as earthquakes, ocean-Earth interactions, atmospheric phenomena, and human-related activities. We address two fundamental, perennial questions in seismology: (a) Does a given seismogram record an earthquake? and (b) Which type of wave motion (e.g., compressional versus shear strain) is predominant in an earthquake seismogram? In Earthquake Science, answering these questions is referred to as~\textit{detecting earthquakes} and~\textit{identifying phases}, respectively. The development of efficient, reliable, and automated solution procedures that can be easily adapted to new environments is critical to modern research and engineering applications in this field, such as in building Earthquake Early Warning Systems. Towards this end, we show that FastMapSVM is a viable ML framework. Through experiments, we show that FastMapSVM's various performance measures, such as precision, recall, and accuracy, are comparable to those of other state-of-the-art methods. However, we also show that, compared to those methods, FastMapSVM uses significantly smaller amounts of time and data for model training. Moreover, FastMapSVM provides a perspicuous visualization of the seismograms, their spread, and the classification boundaries between them.

\par
The key contributions of this paper are as follows:
\begin{enumerate}
\item We advance FastMapSVM to be applicable to domains with complex objects by leveraging a domain-specific distance function to simplify object representation. We propose FastMapSVM as an advantageous alternative to other ML algorithms for general classification tasks.
\item We show how FastMapSVM extends the applicability of SVMs and kernel methods to domains where it is hard to represent the subtle features of individual complex objects but easy to create a distance function on pairs of objects.
\item We illustrate how domain knowledge can be explicitly leveraged into FastMapSVM's reasoning framework via a user-specified distance function that captures $O(N^2)$ pieces of information for $N$ training instances.
\item We demonstrate that FastMapSVM performs comparably to other state-of-the-art methods for classifying seismograms using two orders of magnitude less time and data for model training.
\item We show how FastMapSVM provides a perspicuous visualization of the objects and the classification boundaries between them, aiding human interpretation of the data and results for further decision making.
\item We provide an efficient implementation of FastMapSVM, publicly available at:~\url{https://github.com/malcolmw/FastMapSVM}.
\end{enumerate}

\section*{Results}
\label{sec:results}

\subsection*{Data}
\label{sec:results__data}

\par
We assess the performance and robustness of FastMapSVM using two data sets. All waveforms used in this paper are bandpass filtered between~\SI{1}{\hertz} and~\SI{20}{\hertz} before analysis using a zero-phase Butterworth filter with four poles; we refer to this frequency band as our passband.

\paragraph*{Stanford Earthquake Data Set (STEAD).} The first data set is the Stanford Earthquake Data Set (STEAD)~\cite{mszb19}, a benchmark data set for training and testing algorithms in Earthquake Science. STEAD has been gradually augmented with new data instances since it was first published. The version we use comprises nearly~\SI{1.266e6}{} carefully curated, three-component (3C) seismograms. Data in STEAD contain signals from approximately~\SI{4.417e5}{} different earthquakes\textemdash each recorded by a seismometer located within~\SI{350}{\kilo\meter} of the epicenter\textemdash and represent seismic activity on every continent except Antarctica. About~\SI{2.354e5}{} signals in STEAD comprise only noise (i.e., do not contain earthquake-related signals).

\par
We use the entire STEAD data set to assess model performance for detecting earthquakes and subsets of various sizes (appropriately indicated below) to assess model sensitivity to the size of the training data and the dimensionality of the Euclidean embedding. To assess model performance for identifying phases, we use a subset of~\SI{538}{} three-second, 3C seismograms from STEAD, all of which were recorded by station TA.109C;~\SI{269}{} start~\SI{1}{\second} before a compressional (P-wave) phase arrival, and~\SI{269}{} start~\SI{1}{\second} before a shear (S-wave) phase arrival.

\paragraph*{Ridgecrest Data Set.} The second data set, which we simply refer to as the~\textit{Ridgecrest} data set, comprises data recorded by station CI.CLC of the Southern California Seismic Network (SCSN)~\cite{cu26} on 5 July 2019, the first day of the aftershock sequence following the 2019 Ridgecrest, CA, earthquake pair, and on 5 December 2019, five months after the mainshocks. We use the earthquake catalog published by the Southern California Earthquake Data Center (SCEDC)~\cite{c13} to extract~\SI{512}{} eight-second, 3C seismograms,~\SI{256}{} of which record both P- and S-wave phase arrivals from a nearby aftershock, and the remaining~\SI{256}{} of which record only noise. All~\SI{512}{} of these signals were recorded on 5 July 2019.

\par
We use the Ridgecrest data set to first demonstrate the robustness of FastMapSVM against noisy perturbations. We then use it to demonstrate FastMapSVM's ability to detect new microseisms by automatically scanning a~\SI{10}{\minute}, continuous, 3C seismogram recorded between 01:00:00 and 01:10:00 (UTC) on 5 December 2019. Whereas the analysis on the STEAD demonstrates FastMapSVM's performance on a benchmark, the analysis on the Ridgecrest data set provides an example of a more realistic use case of FastMapSVM: After handpicking only a small number of earthquake and noise signals\textemdash a task that even a novice analyst can perform in a few hours\textemdash continually arriving seismic data can be automatically scanned for additional earthquake signals. This capability manifests the primary conclusion of the preceding robustness test: Even when earthquake signals are difficult to discern by the human eye, FastMapSVM can often reliably detect them.

\subsection*{STEAD Analysis}
\label{sec:result__STEAD_performance}

\paragraph*{Detecting Earthquakes in STEAD.} The~\textit{EQTransformer} DL model~\cite{mezcb20} for simultaneously detecting earthquakes and identifying phase arrivals is arguably the most accurate, publicly available model for this pair of tasks. The authors of EQTransformer report perfect precision and recall scores for detecting earthquakes in~\SI{10}{\percent} of the STEAD waveforms after training its roughly~\SI{3.72e5}{} model parameters with~\SI{85}{\percent} of the STEAD waveforms;~\SI{5}{\percent} of the STEAD waveforms were reserved for model validation. The authors of EQTransformer used a version of the STEAD data set with~\SI{1e6}{} and~\SI{3e5}{} earthquake and noise waveforms, respectively. Their version differs slightly from the newer version we use.

\par
\begin{figure}[!t]
\centering
\includegraphics[width=\textwidth]{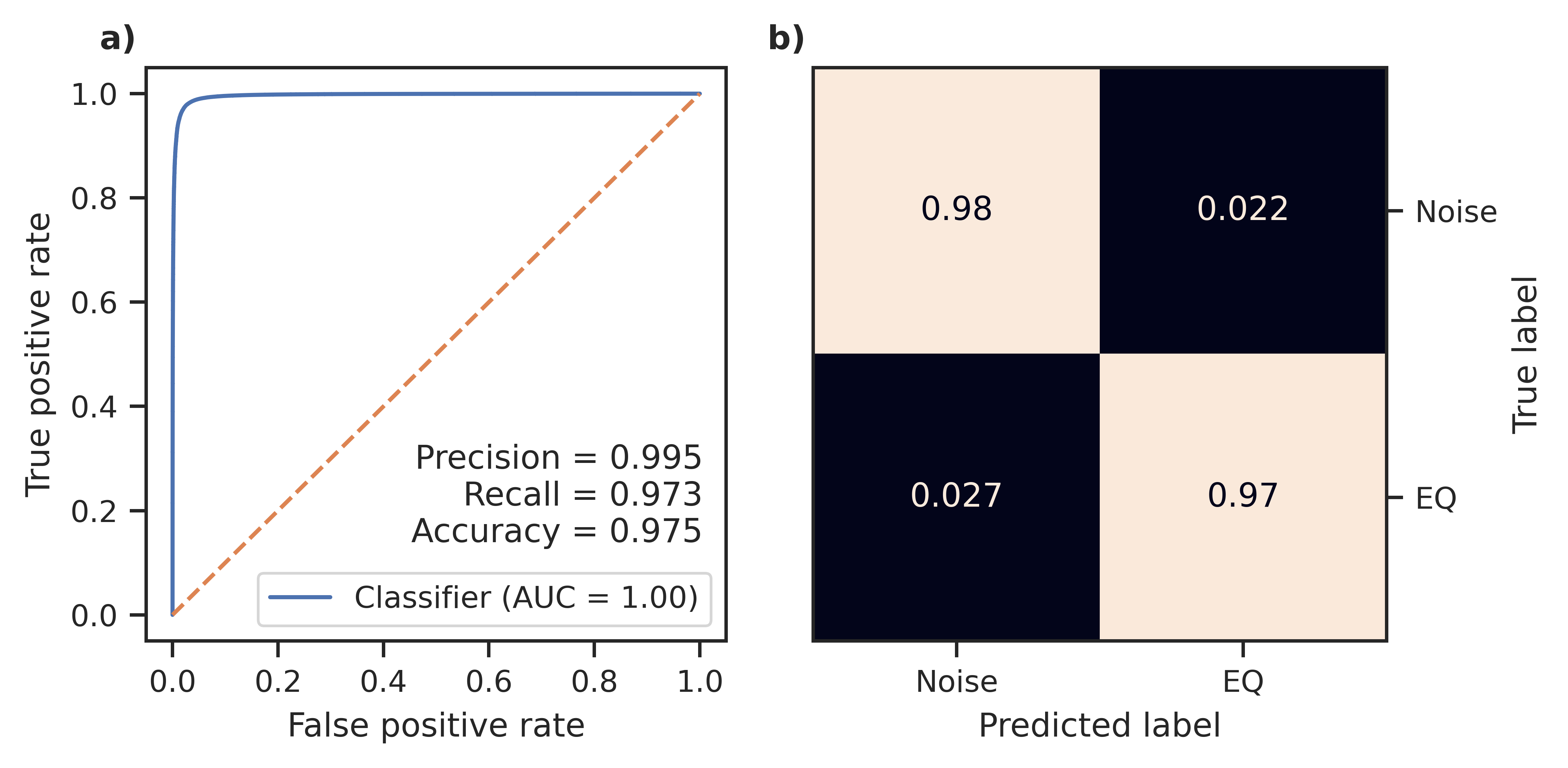}
\caption{\textbf{FastMapSVM's performance on detecting earthquakes in STEAD.} Shows the performance of FastMapSVM on the STEAD data set for classifying Earthquake (EQ) and Noise signals. (a) shows the Receiver Operating Characteristic (ROC) curve and the corresponding Area Under the Curve (AUC). In its inlay, it also shows the precision, recall, and accuracy achieved with the best model parameters. (b) shows the confusion matrix for the learned model with respect to classifying the EQ and Noise signals.}
\label{fig:STEAD_performance}
\end{figure}

Using only $\sim$\SI{1.295}{\percent} (i.e.,~\SI{16384}{}) of the STEAD waveforms, we train FastMapSVM and classify the remaining $\sim$\SI{98.705}{\percent} (i.e., $\sim$\SI{1.253e6}{}) of the STEAD waveforms. FastMapSVM yields the precision, recall, and accuracy scores of~\SI{0.995}{},~\SI{0.973}{}, and~\SI{0.975}{}, respectively. Fig.~\ref{fig:STEAD_performance} and Table~\ref{tbl:STEAD_performance} summarize these performance results. Equal numbers of earthquake and noise waveforms were randomly selected to constitute the training data set, whereas $\sim$\SI{1.022e6}{} earthquake and $\sim$\SI{2.272e5}{} noise waveforms constitute the test data set. FastMapSVM incorrectly labels only $\sim$\SI{2.7}{\percent} of the earthquake waveforms as noise and $\sim$\SI{2.2}{\percent} of the noise waveforms as earthquakes.

\begin{table}[!t]
\caption{\textbf{Model performance comparison.} Shows a comparison between the detection performances of FastMapSVM and other NN models trained on the STEAD data set. All data in this table for the EQTransformer and CRED models are taken from Table 1 of~\textcite{mezcb20}.}
\label{tbl:STEAD_performance}
\centering
\begin{tabular}{lccccc}
\toprule
Model &Precision &Recall &F1 &Training Size &Reference\\
\midrule
EQTransformer &1.0 &1.0  &1.0  &\SI{1.2e6}{} &\textcite{mezcb20}\\
CRED          &1.0 &0.96 &0.98 &\SI{1.2e6}{} &\textcite{mzsb19}\\
FastMapSVM    &1.0 &0.97 &0.98 &\SI{1.638e4}{} &This Article\\
\bottomrule
\end{tabular}
\end{table}

\par
The trained FastMapSVM model uses a 32-dimensional Euclidean embedding of the seismograms, taking~\SI{26}{minutes} to train on a 64-core workstation. This is significantly smaller in comparison to the training requirements of EQTransformer, which takes about~\SI{89}{hours} on 4 parallel Tesla-V100 GPUs~\cite{mezcb20}. FastMapSVM takes about~\SI{5}{hours} for classifying the test data. Using two orders of magnitude less training time and data, FastMapSVM performs competitively against other leading NN models trained to detect earthquakes using the STEAD data set (Table~\ref{tbl:STEAD_performance}). The complexity of the EQTransformer model (and the demands placed on its training time and data) are partly due to the fact that it detects earthquakes and identifies phases simultaneously. Although FastMapSVM can be used for both of these tasks, a separate model must be trained for each. FastMapSVM convincingly outperforms the CRED model~\cite{mzsb19}, which only detects earthquakes and requires the same training data as EQTransformer.

\paragraph*{Sensitivity to Training Data Size and Dimensionality of Euclidean Embedding.} Two important questions concerning FastMapSVM are: (a) How much training data is needed to train the model? and (b) How many Euclidean dimensions are needed to represent the objects being classified? We address both these questions below.

\par
\begin{figure}[!t]
\centering
\includegraphics[width=\textwidth]{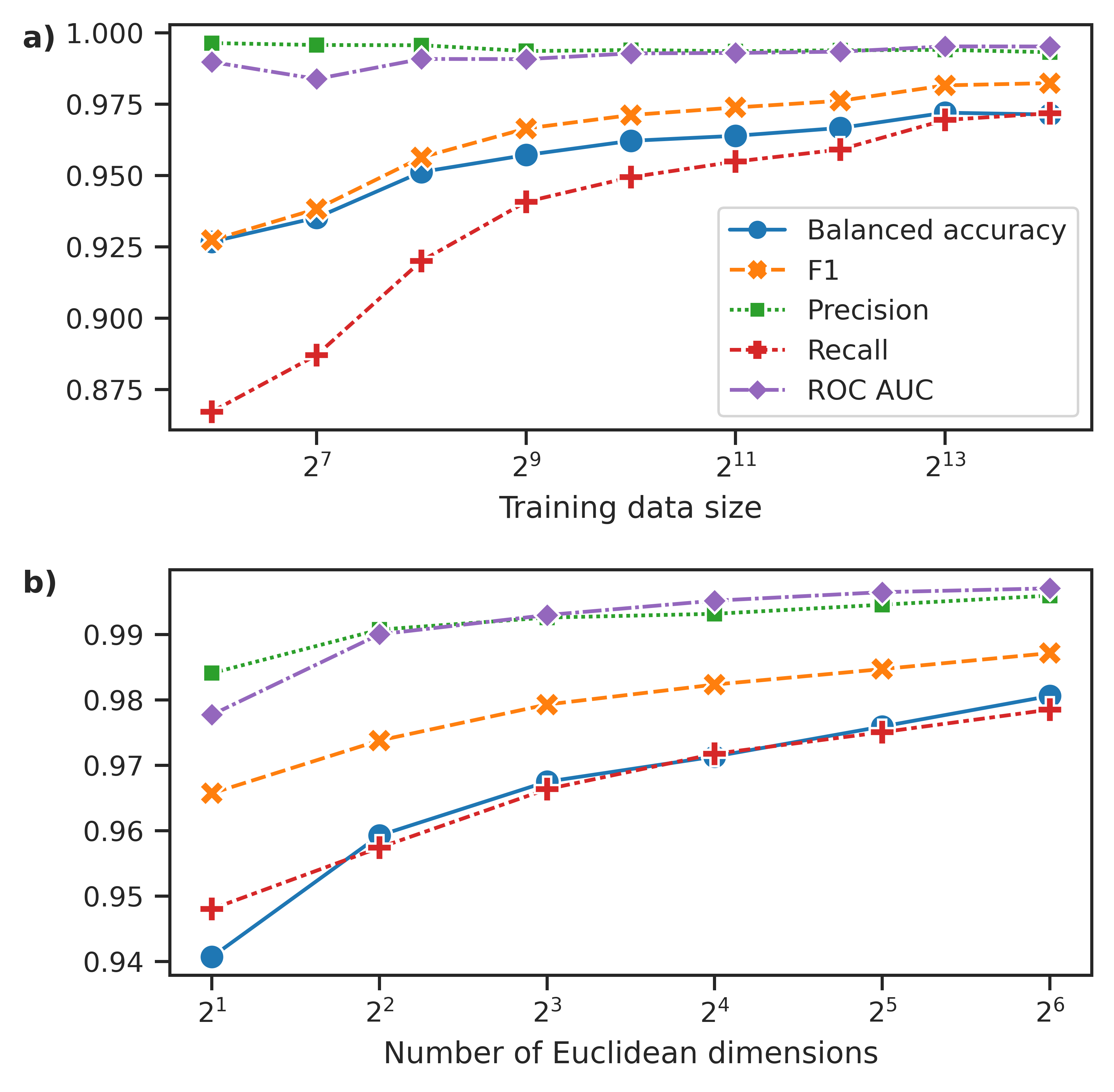}
\caption{\textbf{FastMapSVM's sensitivity to training data size and dimensionality of Euclidean embedding.} Shows the performance of FastMapSVM on the STEAD data set for varying training data size and number of dimensions used for the Euclidean embedding. (a) shows the influence of the training data size, measured using the metrics of balanced accuracy, F1 score, precision, recall, and ROC AUC. (b) shows the influence of the number of dimensions used for the Euclidean embedding, measured using the same metrics.}
\label{fig:performance_breakdown}
\end{figure}

To assess FastMapSVM's sensitivity to the amount of training data used, we obtain a suite of FastMapSVM models trained with various amounts of data (Fig.~\ref{fig:performance_breakdown}a). We score their performances on a subset of~\SI{16384}{} test waveforms randomly selected from STEAD. We ensure that the test data is balanced with equal numbers of earthquake and noise seismograms. The precision appears relatively insensitive to the amount of training data; however, the accuracy and recall increase significantly with the amount of training data. This implies that the FastMapSVM models seldom classify noise as an earthquake, irrespective of the amount of training data. On the other hand, the frequency with which they classify earthquakes as noise decreases as the amount of training data increases. Such behaviour is unsurprising because it is highly unlikely for a noise signal to be more similar to a reference earthquake signal than to a reference noise signal, regardless of how many earthquake signals it is compared to. In contrast, it is relatively more likely for an earthquake signal to be sufficiently dissimilar from all reference earthquake signals and consequently get classified as noise when the number of reference earthquake signals is small. Therefore, generally speaking, correctly identifying noise is an easier task than correctly identifying an earthquake.

\par
To assess FastMapSVM's sensitivity to the dimensionality of the Euclidean embedding, we obtain a suite of FastMapSVM models with a varying number of dimensions (Fig.~\ref{fig:performance_breakdown}b). We score their performances on the same balanced subset of test waveforms used to assess the model sensitivity to training data size above. All performance metrics, particularly the recall, improve with an increasing number of dimensions. Moreover, the performance results are indicative of the ``diminishing returns'' property: Strong performance can be achieved with low-dimensional Euclidean embeddings, although small improvements are possible with high-dimensional Euclidean embeddings. The diminishing returns property is an attractive property from the perspective of visualization in low-dimensional Euclidean spaces and from the perspective of trading off performance against memory.

\begin{figure}[!t]
\centering
\includegraphics[width=\textwidth]{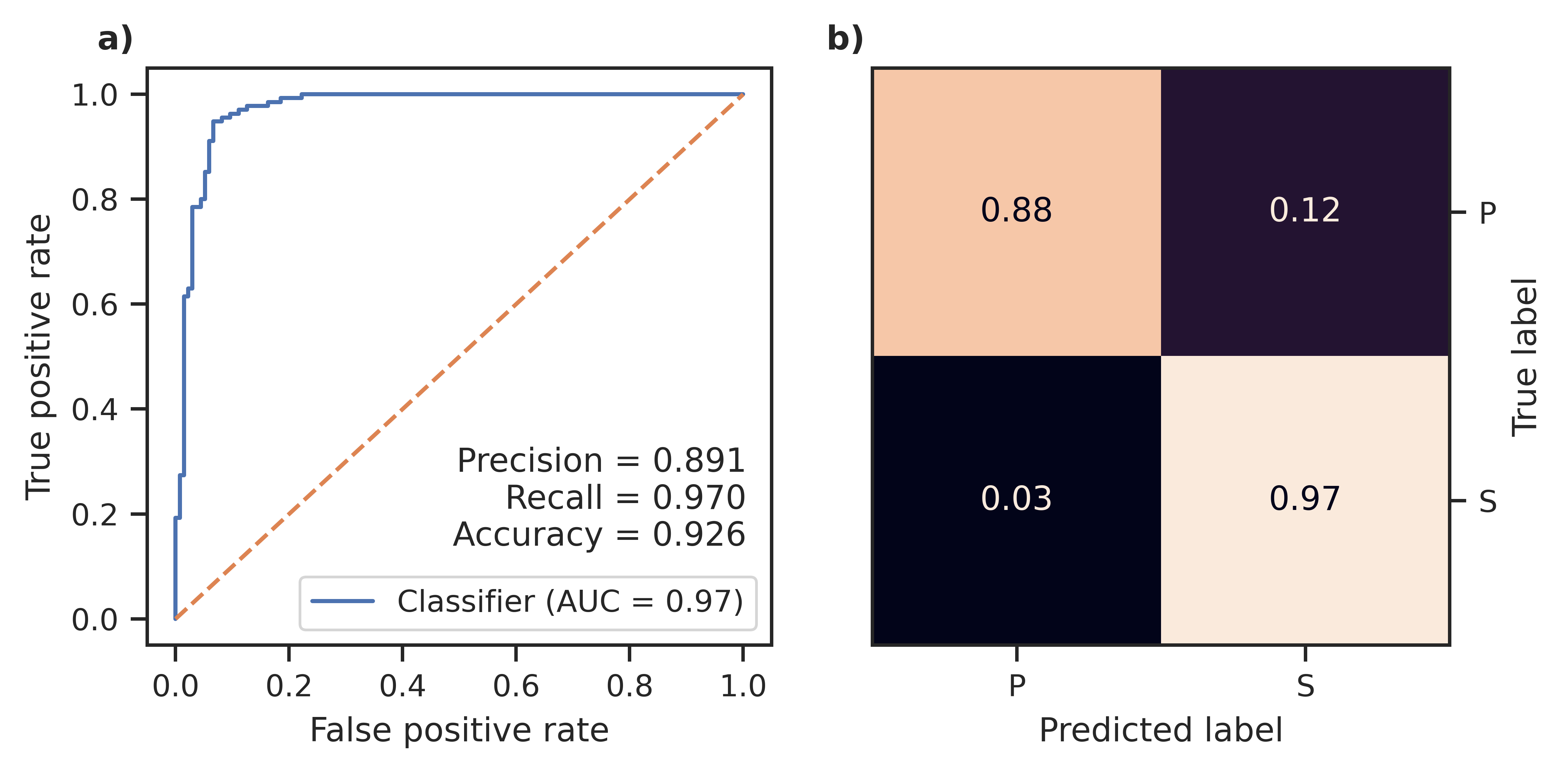}
\caption{\textbf{FastMapSVM's performance on identifying phases for station TA.109C in STEAD.} Shows the performance of FastMapSVM on the STEAD data set for classifying P- and S-waves recorded by station TA.109C. (a) shows the ROC curve and the corresponding AUC. (b) shows the confusion matrix for the learned model with respect to classifying the P- and S-waves.}
\label{fig:TA_109C_PS}
\end{figure}

\paragraph*{Identifying Phase Arrivals.} As another illustration designed to demonstrate the effectiveness of the FastMapSVM framework, we use the STEAD data set to train a model with a 32-dimensional Euclidean embedding (Fig.~\ref{fig:TA_109C_PS}). This model is trained to discriminate P- and S-wave phases using~\SI{268}{} seismograms from STEAD extracted for station TA.109C. We then test the model on~\SI{270}{} seismograms, yielding the precision, recall, and accuracy scores of~\SI{0.891}{},~\SI{0.970}{}, and~\SI{0.926}{}, respectively. The training and test data used in this analysis are both balanced across the P- and S-wave classes. Although these scores are relatively modest in comparison to those of state-of-the-art NNs designed for similar tasks, they demonstrate that FastMapSVM can be easily trained for strong performance using only small amounts of time and data.

\subsection*{Ridgecrest Analysis}

\begin{figure}[!t]
\centering
\includegraphics[width=\textwidth]{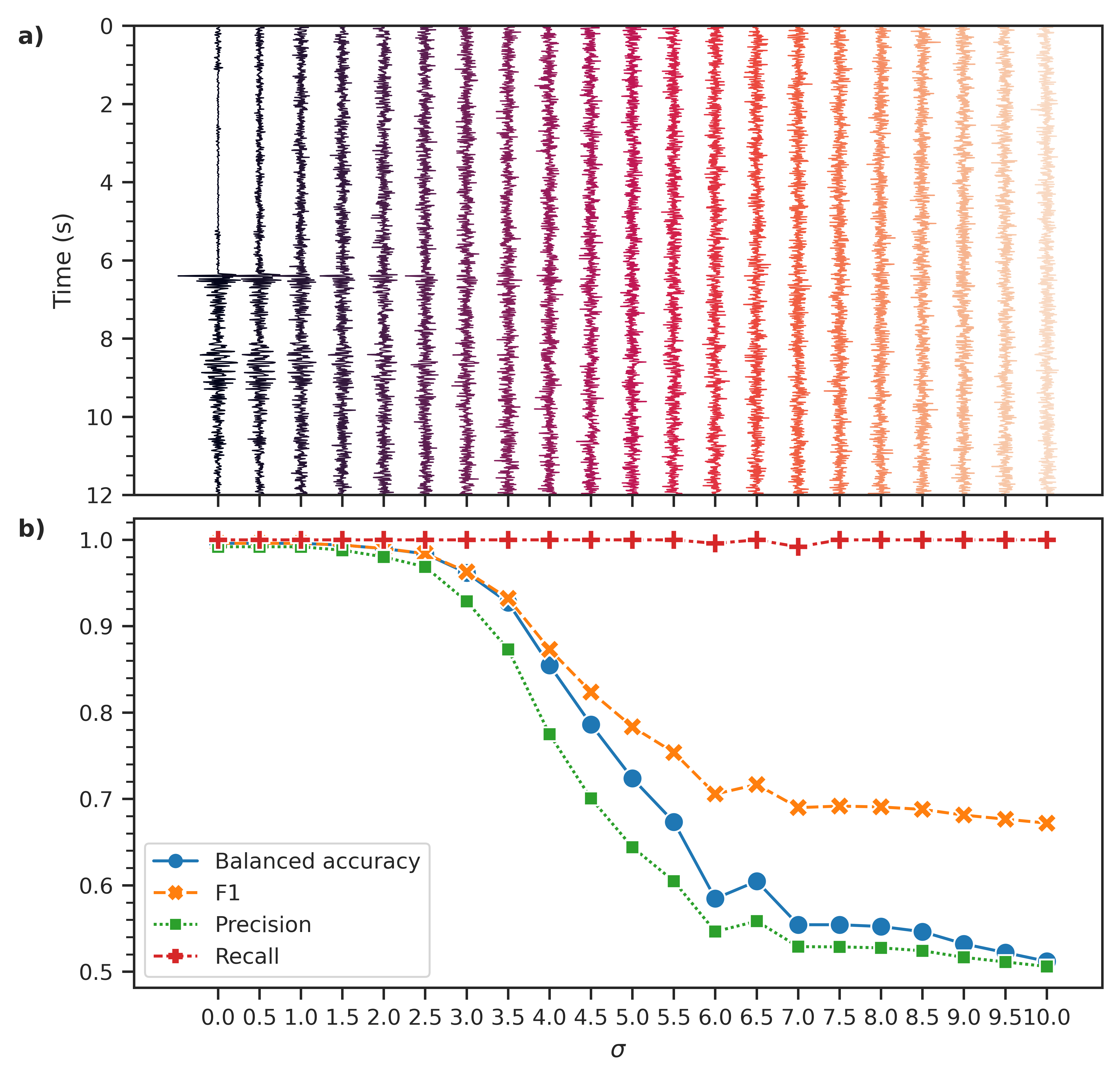}
\caption{\textbf{FastMapSVM's robustness against noisy perturbations.} Shows the performance of FastMapSVM on the Ridgecrest data set. (a) shows how a sample test waveform changes with the addition of increasing levels of Gaussian random noise with mean $0$ and standard deviation $\sigma$. It uses a vertical time-axis and an increasing $\sigma$ on the horizontal axis. (b) shows how the metrics of balanced accuracy, F1 score, precision, and recall change with increasing $\sigma$.}
\label{fig:noise_resiliency}
\end{figure}

\paragraph*{Robustness against Noisy Perturbations.} It is critical that a classification framework is robust against noisy perturbations of inputs. In general, the robustness of FastMapSVM against noisy perturbations may depend on the characteristics of the data and the chosen distance function. For classifying seismograms, we demonstrate FastMapSVM's robustness against noisy perturbations made to the Ridgecrest data set using the distance function described in the Materials and Method section. We randomly select~\SI{8}{} earthquake signals and~\SI{8}{} noise signals to train a FastMapSVM model with a 4-dimensional Euclidean embedding. Each of the~\SI{496}{} remaining seismograms is used as a test instance. Each test instance is circularly shifted by an offset (in seconds) chosen uniformly at random from the interval $[-2, 2]$. FastMapSVM has a nearly perfect classification accuracy; only~\SI{2}{} noise signals are incorrectly labeled as earthquakes. We subsequently conduct a set of experiments in which this model's performance is scored after perturbing signals in the test data set with increasing amounts of Gaussian noise. For each trial, we perturb each signal in the test data set by adding Gaussian noise with mean $0$ and standard deviation $\sigma$; $\sigma$ increases by $0.5$ after each trial. Fig.~\ref{fig:noise_resiliency}a shows how a waveform changes with increasing $\sigma$. Fig.~\ref{fig:noise_resiliency}b shows the performance of FastMapSVM with increasing $\sigma$. We observe that FastMapSVM continues to classify seismograms with high fidelity, even as earthquake signals become indiscernible to the human eye; e.g., the FastMapSVM model achieves $>$\SI{90}{\percent} precision and accuracy for $\sigma = 3$.

\par
At first glance, some of the results of the foregoing experiments are counterintuitive. The recall remains at or close to~\SI{1}{} irrespective of the amplitude of the noisy perturbations: The model accurately identifies earthquakes regardless of the magnitude of the noisy perturbations. In fact, the model misclassifies noise signals as earthquake signals more frequently when the magnitude of the noisy perturbations is increased. With enough added noise, the model classifies all signals as earthquake signals. This is because of the unique frequency content of the noisy perturbations (Supplementary Fig.~\ref{fig:suppl_frequency_spectra}). In our passband, the average frequency spectrum of earthquake signals is nearly flat; whereas the average frequency spectrum of noise signals has prominent peaks near the low- and high-frequency endpoints. Because the noisy perturbations are Gaussian, their frequency spectrum is flat. This makes the frequency spectra of noisy perturbations more similar to those of earthquake signals than those of real noise signals. Thus, the recall and F1 scores get inflated when the amount of added noise increases. However, the precision and accuracy remain unbiased because precision penalizes false positives in equal proportion to rewarding true positives and accuracy is not directly sensitive to false positives.

\begin{figure}[!t]
\centering
\includegraphics[width=\textwidth]{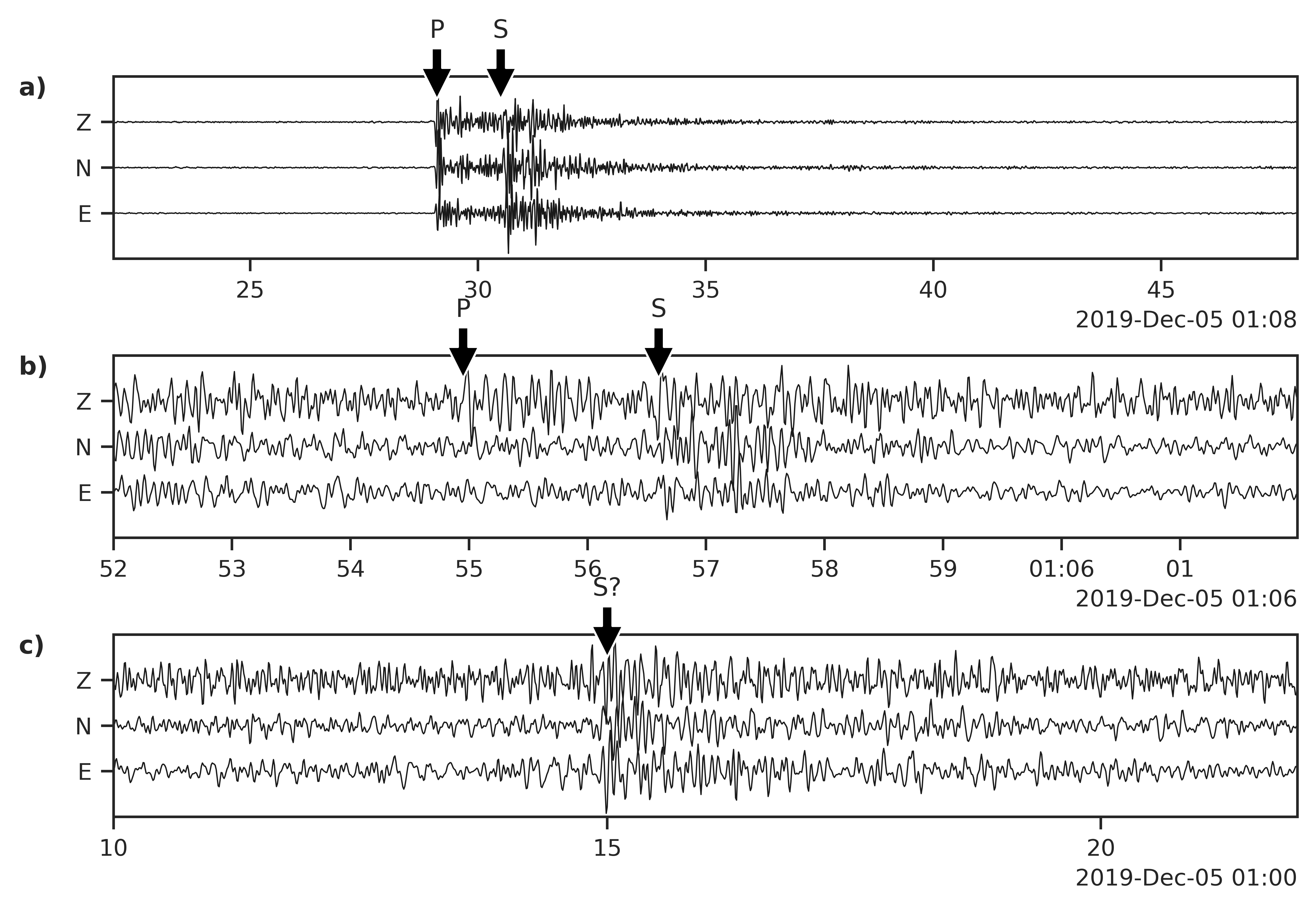}
\caption{\textbf{Example results from an automatic scan for earthquake signals using FastMapSVM.} Shows example results of automatically scanning~\SI{600}{\second} of data recorded by station CI.CLC. (a) shows a clear earthquake signal with easily discernible P- and S-wave arrivals. (b) shows an earthquake signal with low signal-to-noise ratio. The P- and S-wave arrivals are close to the noise level. (c) shows an ambiguous signal that may or may not be from an earthquake source.}
\label{fig:blind_scan}
\end{figure}

\paragraph*{Automatic Scanning.} We further demonstrate a use case-inspired application of FastMapSVM. We first train a model with~\SI{128}{} earthquake signals and~\SI{128}{} noise signals selected randomly from the Ridgecrest data set. We then use the trained model to automatically scan and detect earthquakes in a~\SI{10}{\minute}, continuous seismogram recorded by station CI.CLC between 01:00:00 and 01:10:00 (UTC) on 5 December 2019. We validate the results after automatically scanning the data. During this time period, the SCEDC earthquake catalog reports no earthquakes within~\SI{100}{\kilo\meter} of CI.CLC; however, FastMapSVM identifies~\SI{19}{} windows with earthquakes. Of these,~\SI{9}{} contain clear earthquake signals with easily discernible P- and S-wave arrivals (Fig.~\ref{fig:blind_scan}a). Another~\SI{7}{} of them contain signals that we believe are from earthquake sources but are difficult to discern, either because they have low signal-to-noise ratios, secondary phase arrivals, or both (Fig.~\ref{fig:blind_scan}b). The remaining~\SI{3} of them have ambiguous signals that may or may not be from genuine earthquake sources (Fig.~\ref{fig:blind_scan}c). The complete set of waveforms identified as containing earthquakes in this test, along with our manual categorizations of them, are available in the Supplementary Material (Figs.~\ref{fig:suppl_clear},~\ref{fig:suppl_low_snr}, and~\ref{fig:suppl_ambiguous}).

\section*{Discussion}
\label{sec:discussion}

\par
Although FastMapSVM was first developed by~\textcite{bka09}, our conception and advancement of it for classifying complex objects is independent work. The work of~\textcite{bka09} only emphasized the sparse kernel representations enabled by FastMap. Their experiments focused on analyzing the performance of SVMs with such sparse kernel representations in comparison to regular SVMs. In contrast, our work focuses on the applicability of FastMapSVM in real-world domains where the objects are overwhelmingly complex for regular SVMs to be effective. In such cases, we leverage a domain-specific distance function on pairs of objects: It is often easier to build compared to identifying the subtle features of individual complex objects. Intuitively, FastMap is used as a ``representational counterpart'' to SVMs in FastMapSVM. Therefore, we focus on manifesting the benefits of FastMapSVM in complex domains where a combination of representation and classification learning is required. Moreover, we compare FastMapSVM against other ML methods, such as NNs, that seamlessly integrate representation and classification learning. We show that FastMapSVM has many advantages over existing ML methods for classifying complex objects such as seismograms. These aspects and advantages of FastMapSVM were largely overlooked by~\textcite{bka09}. In this section, we discuss some of the advantages of FastMapSVM, both in the specific context of classifying seismograms and in the general context of ML and data visualization.

\par
Many existing ML algorithms for classification do not leverage domain knowledge when used off the shelf. Although a domain expert can occasionally incorporate domain-specific features of the objects being classified into the classification task, doing so becomes increasingly difficult as the complexity of the objects increases. FastMapSVM enables domain experts to incorporate their domain knowledge via a distance function instead of relying on complex ML models to infer the underlying structure in the data entirely. In fact, in many real-world domains, it is easier to construct a distance function on pairs of objects than it is to extract features of individual objects. Examples include DNA strings, for which the edit distance is well defined, images, for which the Minkowski distance~\cite{achh16} is well defined, and text documents, for which the cosine similarity~\cite{rka12} is well defined. In all these domains, extracting features of individual objects is challenging. In the seismogram domain, our~\textit{a priori} knowledge that earthquake seismograms typically bear similarities to one another is encapsulated in a distance function that quantifies the normalized cross-correlation of the waveforms. This distance metric closely resembles other similarity metrics that have been extensively used in previous works in the Earthquake Science community~\cite{gr06,sbi07,seh16}.

\par
\begin{figure}[!t]
\centering
\includegraphics[width=\textwidth]{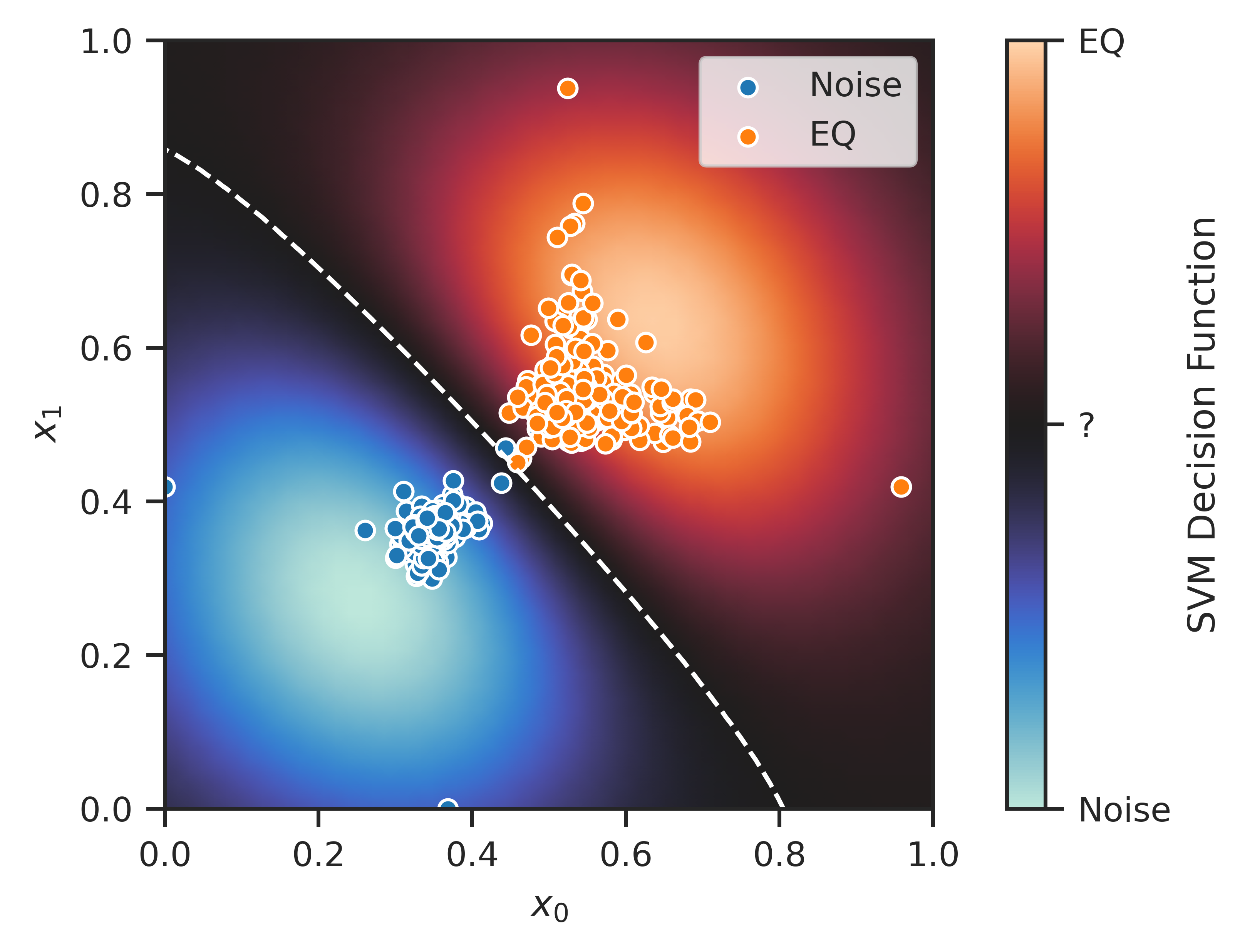}
\caption{\textbf{Perspicuous visualization of seismograms and classification boundaries produced by FastMapSVM.} Shows a visualization of FastMapSVM's classification boundary (dashed, white curve) and decision function (background) in a 2-dimensional Euclidean embedding of the training data from the Ridgecrest data set. EQ refers to earthquakes.}
\label{fig:embedding}
\end{figure}

In addition, many existing ML algorithms produce results that are hard to interpret or explain. For example, in NNs, a large number of interactions between neurons with nonlinear activation functions makes a meaningful interpretation or explanation of the results challenging. In fact, the very complexity of the objects in the domain can hinder interpretability and explainability. FastMapSVM mitigates these challenges and thereby supports interpretability and explainability. Although the objects themselves may be complex, FastMapSVM embeds them in a Euclidean space by considering only the distance function defined on pairs of objects. In effect, it simplifies the description of the objects by assigning Euclidean coordinates to them. Moreover, because the distance function is itself user-supplied and encapsulates domain knowledge, FastMapSVM naturally facilitates interpretability and explainability. It even provides a perspicuous visualization of the objects and the classification boundaries between them (Fig.~\ref{fig:embedding}). FastMapSVM produces such visualizations very efficiently because it invests only linear time in generating the Euclidean embedding.

\par
FastMapSVM also uses significantly smaller amounts of time and data for model training compared to other ML algorithms. While NNs and other ML algorithms store abstract representations of the training data in their model parameters, FastMapSVM stores explicit references to some of the original objects, referred to as pivots. While making predictions, test instances are compared directly to the pivots in the original data domain using the user-supplied distance function. FastMapSVM thereby obviates the need to learn a complex transformation of the input data and thus significantly reduces the amount of time and data required for model training. Moreover, given $N$ training instances, FastMapSVM leverages $O(N^2)$ pieces of information via the distance function, which is defined on every pair of objects. In contrast, ML algorithms that focus on individual objects leverage only $O(N)$ pieces of information.

\par
In general, FastMapSVM extends the applicability of SVMs and kernel methods to domains with complex objects. With increasing complexity of the objects, deep NNs have gained more popularity compared to SVMs because it is unwieldy for SVMs to represent all the features of complex objects in Euclidean space. FastMapSVM, however, revitalizes the SVM approach by leveraging a distance function and creating a low-dimensional Euclidean embedding of the objects.

\par
Overall, any application domain hindered by a paucity of training data but possessing a well-defined distance function on pairs of its objects can benefit from the advantages of FastMapSVM. Examples of such applications in Earthquake Science include analyzing and learning from data obtained by distributed acoustic sensing technology or during temporary deployments of ``large-$N$'' nodal arrays. Furthermore, the efficiency of FastMapSVM makes it suitable for real-time deployment, which is critical for engineering Earthquake Early Warning Systems.

\section*{Materials and Method}
\label{sec:method}

\par
Our FastMapSVM method comprises two main components: (a) The FastMap algorithm~\cite{fl95} for embedding complex objects in a Euclidean space using a distance function, and (b) SVMs for classifying objects in the resulting Euclidean space. We explain the key algorithmic concepts behind each of these components below.

\paragraph*{Review of the FastMap Algorithm.} FastMap~\cite{fl95} is a Data Mining algorithm that embeds complex objects\textemdash such as audio signals, seismograms, DNA sequences, electrocardiograms, or magnetic-resonance images\textemdash into a $K$-dimensional Euclidean space, for a user-specified value of $K$ and a user-supplied function $\mathcal{D}$ that quantifies the distance, or dissimilarity, between pairs of objects. The Euclidean distance between any two objects in the embedding produced by FastMap approximates the domain-specific distance between them. Therefore, similar objects, as quantified by $\mathcal{D}$, map to nearby points in Euclidean space whereas dissimilar objects map to distant points. Although FastMap preserves $O(N^2)$ pairwise distances between $N$ objects, it generates the embedding in only $O(KN)$ time. Because of its efficiency, FastMap has already found numerous real-world applications, including in Data Mining~\cite{fl95}, shortest-path computations~\cite{cujakk18}, community detection and block modeling~\cite{lskk22}, and solving combinatorial optimization problems on graphs~\cite{lfkk19}.

\par
Below, we review the FastMap algorithm~\cite{fl95} and describe our minor modifications to it. These modifications suit the purposes of the downstream classification task. Our review of FastMap also serves completeness and the readers' convenience.

\par
FastMap embeds a collection of complex objects in an artificially created Euclidean space that enables geometric interpretations, algebraic manipulations, and downstream application of ML algorithms. It gets as input a collection of complex objects $\mathcal{O}$ and a distance function $\mathcal{D}(\cdot, \cdot)$, where $\mathcal{D}(O_i, O_j)$ represents the domain-specific distance between objects $O_i, O_j \in \mathcal{O}$. It generates a Euclidean embedding that assigns a $K$-dimensional point $\mathbf{p}_i = \left(p_{i,1}, p_{i,2}, \ldots, p_{i,K}\right) \in \mathbb{R}^K$ to each object $O_i$. A good Euclidean embedding is one in which the Euclidean distance $\lVert \mathbf{p}_i - \mathbf{p}_j \rVert_2 \equiv \sqrt{\sum_{n = 1}^K(p_{i,n} - p_{j,n})^2}$ between any two points $\mathbf{p}_i$ and $\mathbf{p}_j$ closely approximates $\mathcal{D}(O_i, O_j)$.

\par
\begin{figure}[!t]
\centering
\includegraphics[width=\textwidth]{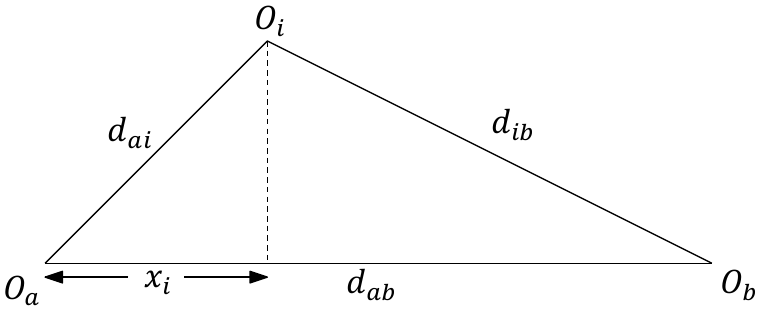}
\caption{\textbf{``Cosine law'' employed by FastMap.} The ``cosine law'' projection in a triangle.}
\label{fig:projection_1d}
\end{figure}

FastMap creates a $K$-dimensional Euclidean embedding of the complex objects in $\mathcal{O}$, for a user-specified value of $K$. In the first iteration, FastMap heuristically identifies the farthest pair of objects $O_a$ and $O_b$ in linear time. Once $O_a$ and $O_b$ are determined, every other object $O_i$ defines a triangle with sides of lengths $d_{ai} = \mathcal{D}(O_a,O_i)$, $d_{ab} = \mathcal{D}(O_a,O_b)$, and $d_{ib} = \mathcal{D}(O_i,O_b)$ (Fig.~\ref{fig:projection_1d}). The sides of the triangle define its entire geometry, and the projection of $O_i$ onto the line $\overline{O_aO_b}$ is given by
\begin{equation}
x_i = (d_{ai}^2 + d_{ab}^2 - d_{ib}^2) / (2d_{ab}).
\label{eqn:projection_1d}
\end{equation}

\par
\begin{figure}[!t]
\centering
\includegraphics[width=\textwidth]{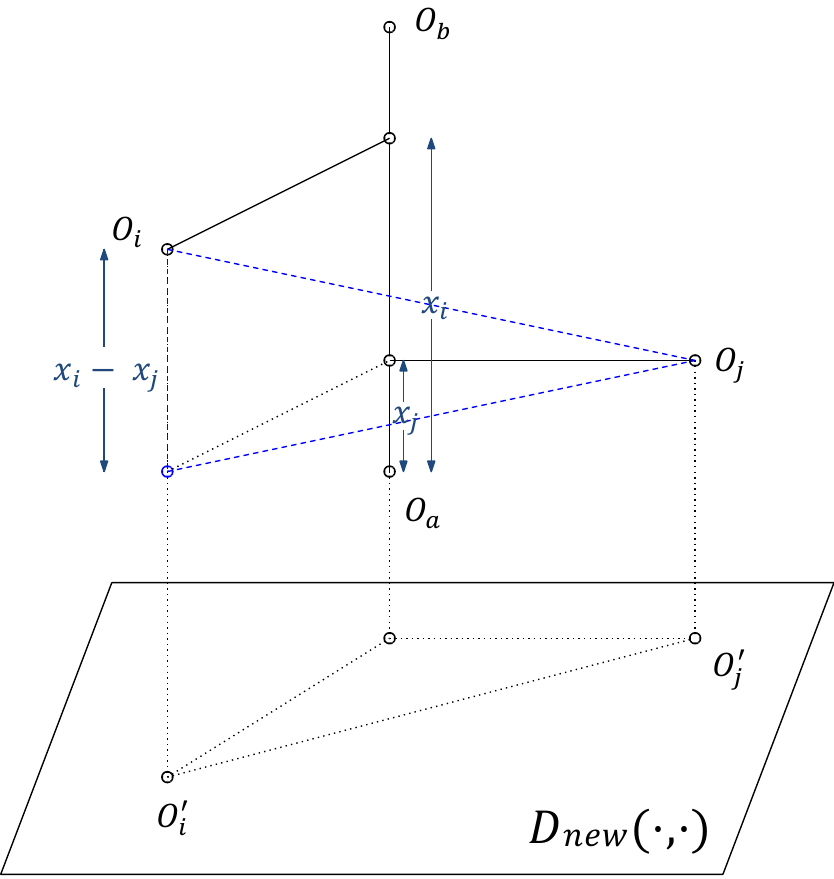}
\caption{\textbf{Hyperplane projection employed conceptually by FastMap.} Projection onto a hyperplane that is perpendicular to $\overline{O_aO_b}$.}
\label{fig:hyperplane}
\end{figure}

FastMap sets the first coordinate of $\mathbf{p}_i$, the embedding of $O_i$, equal to $x_i$. In the subsequent $K-1$ iterations, FastMap computes the remaining $K-1$ coordinates of each object following the same procedure; however, the distance function is adapted for each iteration. In the first iteration, the coordinates of $O_a$ and $O_b$ are $0$ and $d_{ab}$, respectively. Because these coordinates perfectly encode the true distance between $O_a$ and $O_b$, the rest of $\mathbf{p}_a$ and $\mathbf{p}_b$'s coordinates should be identical for all subsequent iterations. Intuitively, this means that the second iteration should mimic the first one on a hyperplane that is perpendicular to the line $\overline{O_aO_b}$ (Fig.~\ref{fig:hyperplane}). Although the hyperplane is never explicitly constructed, it conceptually implies that the distance function for the second iteration should be changed for all $i$ and $j$ in the following way:
\begin{equation}
\mathcal{D}_{new}(O'_i, O'_j)^2 = \mathcal{D}(O_i, O_j)^2 - (x_i - x_j)^2,
\label{eqn:updated_distance}
\end{equation}
\noindent in which $O'_i$ and $O'_j$ are the projections of $O_i$ and $O_j$, respectively, onto this hyperplane, and $D_{new}(\cdot,\cdot)$ is the new distance function. The distance function is recursively updated according to Equation~\ref{eqn:updated_distance} at the beginning of each of the $K-1$ iterations that follow the first.

\paragraph*{Selecting Reference Objects.} As described before, in each of the $K$ iterations, FastMap heuristically finds the farthest pair of objects according to the distance function defined for that iteration. These objects are called pivots and are stored as reference objects. There are exactly $2K$ reference objects in our implementation because we prohibit any object from serving as a reference object more than once; however, this restriction is not strictly necessary. Technically, finding the farthest pair of objects in any iteration takes $O(N^2)$ time. However, FastMap uses a linear-time ``pivot changing'' heuristic~\cite{fl95} to efficiently and effectively identify a pair of objects $O_a$ and $O_b$ that is very often the farthest pair. It does this by initially choosing a random object $O_b$ and then choosing $O_a$ to be the farthest object away from $O_b$. It then reassigns $O_b$ to be the farthest object away from $O_a$.

\par
In our adaptation of FastMap as a component of FastMapSVM, we require the farthest pair of objects $O_a$ and $O_b$ in each iteration to be of opposite classes. This maximizes the discriminatory power of the downstream SVM classifier. We achieve this requirement by implementing a minor modification of the pivot changing heuristic: We initially choose a random object $O_b$. We then choose $O_a$ to be the farthest object away from $O_b$ and of the opposite class. We finally reassign $O_b$ to be the farthest object away from $O_a$ and of the opposite class. In each iteration, all previously used reference objects are excluded from consideration when selecting the pivots.

\par
For a test object not seen before, its Euclidean coordinates in the $K$-dimensional embedding can be computed by using only its distances to the reference objects. This is based on the reasonable assumption that the new test object would not preclude the stored reference objects from being pivots if the $K$-dimensional Euclidean embedding was recomputed along with the new test object. In any case, the assumption is not strictly required since the stored reference objects are close to being the farthest pairs.

\paragraph*{Choosing the Distance Function $\mathcal{D}$.} The distance function should yield non-negative values for all pairs of objects and $0$ for identical objects. We can use a variety of distance functions, such as the Wasserstein distance, the Jensen-Shannon divergence, or the Kullback-Leibler divergence. We can also use more domain-specific knowledge in the distance function, as described below.

\par
In the Earthquake Science community, the normalized cross-correlation operator, denoted here by $\star$, is popularly used to measure similarity between two waveforms. For two zero-mean, single-component seismograms $O_i$ and $O_j$ with lengths $n_i$ and $n_j$, respectively, and starting with index $0$, the normalized cross-correlation is defined with respect to a lag $\tau$ as follows:
\begin{align}
(O_i \star O_j)[\tau] & \triangleq \frac{1}{\sigma_i \sigma_j} \sum_{m = 0}^{n_i-1} O_i[m]\widehat{O}_j[m + \ell - \tau],
\label{eqn:cross_correlation_operator}
\end{align}
\noindent in which, without loss of generality, we assume that $n_i \ge n_j$. $\sigma_i$ and $\sigma_j$ are the standard deviations of $O_i$ and $O_j$, respectively. Moreover, $\ell$ and $\widehat{O}_j$ are defined as follows:
\begin{equation}
\ell \triangleq \frac{n_j-n_j\Mod{2}}{2} - \left(n_i\Mod{2}\right)\left(1-n_j\Mod{2}\right)
\end{equation}
\noindent and
\begin{equation}
\widehat{O}_j[m] \triangleq
\begin{cases}
O_j[m] & \text{if}~0 \le m < n_j\\
0 & \text{otherwise}
\end{cases}.
\end{equation}

\par
Equipped with this knowledge, we first define the following distance function that is appropriate for waveforms with a single component:
\begin{align}
\mathcal{D}(O_i, O_j) & \triangleq 1 - \max_{0 \leq \tau \leq n_i - 1}\left|(O_i \star O_j)[\tau]\right|.
\end{align}
\noindent Based on this, we define the following distance function that is appropriate for waveforms with $L$ components:
\begin{align}
\mathcal{D}(O_i, O_j) & \triangleq 1 - \frac{1}{L} \max_{0 \leq \tau \leq n_i - 1} \left|\sum_{l = 1}^L (O^l_i \star O^l_j)[\tau]\right|.
\label{eqn:correlation_distance}
\end{align}
\noindent Here, each component $O^l_i$ of $O_i$, or $O^l_j$ of $O_j$, is a channel representing a 1-dimensional data stream. A channel is associated with a single standalone sensor or a single sensor in a multi-sensor array.

\par
We use the distance function defined in Equation~\ref{eqn:correlation_distance} with $L = 3$ for 3C seismograms. Our choice is motivated by the extensive use of similar equations in Earthquake Science to detect earthquakes using matched filters~\cite{gr06,sbi07,seh16}. We will investigate other distance functions in future work.

\paragraph*{Enabling SVMs and Kernel Methods.} SVMs are particularly good for classification tasks. When combined with kernel functions, they recognize and represent complex nonlinear classification boundaries very elegantly~\cite{s03}. Moreover, soft-margin SVMs with kernel functions~\cite{pc13} can be used to recognize both outliers and inherent nonlinearities in the data. While the SVM machinery is very effective, it requires the objects in the classification task to be represented as points in a Euclidean space. Often, it is very difficult to represent complex objects like seismograms as precise geometric points without introducing inaccuracy or losing domain-specific representational features. In such cases, NNs have been more effective than SVMs. FastMapSVM revitalizes the SVM technology for classifying complex objects by leveraging the following observation: Although it may be hard to precisely describe complex objects as geometric points, it is often relatively easy to precisely compute the distance between any two of them. FastMapSVM uses the distance function to construct a low-dimensional Euclidean embedding of the objects. It then invokes the full power of SVMs. The low-dimensional Euclidean embedding also facilitates a perspicuous visualization of the classification boundaries.

\paragraph*{Implementing FastMapSVM.} We have implemented FastMapSVM and have made it publicly accessible in a Python package available at:~\url{https://github.com/malcolmw/FastMapSVM}. The most expensive computations, i.e., evaluations of the distance function, are parallelized using Python's built-in~\texttt{multiprocessing} module, which allows for the concurrent execution of multiple threads on a single host. FastMapSVM requires as input (a) the labeled training data set, (b) the distance function, and (c) a location to store the resulting trained model. We used the~\texttt{scikit-learn} SVM implementation and conducted a grid search for the optimal SVM hyperparameters.

\section*{Conclusions and Future Work}
\label{sec:conclusions}

\par
In this paper, we advance FastMapSVM\textemdash an interpretable ML framework that combines the complementary strengths of FastMap and SVMs\textemdash as an advantageous alternative to existing methods, such as NNs, for classifying complex objects. FastMapSVM offers several advantages. First, it enables domain experts to incorporate their domain knowledge using a distance function. This avoids relying on complex ML models to infer the underlying structure in the data entirely. Second, because the distance function encapsulates domain knowledge, FastMapSVM naturally facilitates interpretability and explainability. In fact, it even provides a perspicuous visualization of the objects and the classification boundaries between them. Third, FastMapSVM uses significantly smaller amounts of time and data for model training compared to other ML algorithms. Fourth, it extends the applicability of SVMs and kernel methods to domains with complex objects.

\par
We demonstrated the efficiency and effectiveness of FastMapSVM in the context of classifying seismograms. On the STEAD data set, we showed that FastMapSVM performs comparably to state-of-the-art NN models in terms of precision, recall, and accuracy. It also uses significantly smaller amounts of time and data for model training compared to other methods. On the Ridgecrest data set, we first demonstrated the robustness of FastMapSVM against noisy perturbations. We then demonstrated its ability to reliably detect new microseisms that are otherwise difficult to detect.

\par
In future work, we expect FastMapSVM to be viable for classification tasks in many other real-world domains. In Earthquake Science, we will apply FastMapSVM to analyze and learn from data obtained during temporary deployments of large-$N$ nodal arrays and distributed acoustic sensing. In Computational Astrophysics, we anticipate the use of FastMapSVM for identifying galaxy clusters based on cosmological observations. In general, the efficiency and effectiveness of FastMapSVM also make it suitable for real-time deployment in dynamic environments in applications such as Earthquake Early Warning Systems.

\par
Our implementation of FastMapSVM is publicly available at:~\url{https://github.com/malcolmw/FastMapSVM}.

\printbibliography

@inproceedings{lskk22,
author={Li, Ang and Stuckey, Peter and Koenig, Sven and Kumar, T.~K.~Satish},
booktitle={Proceedings of the International Conference on the Integration of Constraint Programming, Artificial Intelligence, and Operations Research},
title={A {FastMap}-based algorithm for block modeling},
year={2022}
}

@inproceedings{bka09,
author={Ban, Tao and Kadobayashi, Youki and Abe, Shigeo},
booktitle={Proceedings of the International Joint Conference on Neural Networks},
number={1},
pages={256--263},
title={Sparse kernel feature analysis using {FastMap} and its variants},
year={2009}
}

@inproceedings{cujakk18,
author={Cohen, Liron and Uras, Tansel and Jahangiri, Shiva and Arunasalam, Aliyah and Koenig, Sven and Kumar, T.~K.~Satish},
title={The {FastMap} algorithm for shortest path computations},
booktitle={Proceedings of the International Joint Conference on Artificial Intelligence},
year={2018}
}

@inproceedings{lfkk19,
author={Li, Jiaoyang and Felner, Ariel and Koenig, Sven and Kumar, T.~K.~Satish},
title={Using {FastMap} to solve graph problems in a {Euclidean} space},
booktitle={Proceedings of the International Conference on Automated Planning and Scheduling},
year={2019}
}

@article{s03,
author={S{\'a}nchez A, V David},
title={Advanced support vector machines and kernel methods},
journal={Neurocomputing},
year={2003}
}

@inproceedings{pc13,
author={Patle, Arti and Chouhan, Deepak Singh},
title={{SVM} kernel functions for classification},
booktitle={Proceedings of the International Conference on Advances in Technology and Engineering},
year={2013}
}

@article{achh16,
author={Ali, Abder-Rahman and Couceiro, Micael S and Hassanien, Aboul Ella and Hemanth, D Jude},
title={Fuzzy c-means based on {Minkowski} distance for liver {CT} image segmentation},
journal={Intelligent Decision Technologies},
year={2016}
}

@inproceedings{rka12,
author={Rahutomo, Faisal and Kitasuka, Teruaki and Aritsugi, Masayoshi},
title={Semantic cosine similarity},
booktitle={Proceedings of the International Student Conference on Advanced Science and Technology},
year={2012}
}

@inproceedings{hlvw17,
author={Huang, Gao and Liu, Zhuang and Van Der Maaten, Laurens and Weinberger, Kilian Q},
title={Densely connected convolutional networks},
booktitle={Proceedings of the IEEE Conference on Computer Vision and Pattern Recognition},
year={2017}
}

@article{entfhmwpn20,
author={Elgendi, Mohamed and Nasir, Muhammad Umer and Tang, Qunfeng and Fletcher, Richard Ribon and Howard, Newton and Menon, Carlo and Ward, Rabab and Parker, William and Nicolaou, Savvas},
title={The performance of deep neural networks in differentiating chest {X-rays} of {COVID-19} patients from other bacterial and viral pneumonias},
journal={Frontiers in Medicine},
year={2020}
}

@techreport{lhcawnt19,
author={Lin, Zhen and Huang, Nicholas and Caldeira, Joao and Avestruz, Camille and Wu, Kimmy and Nord, Brian and Trivedi, Shubhendu},
title={Classification of {Sunyaev-Zel'dovich} galaxy clusters using deep learning},
institution={Fermi National Accelerator Lab, Batavia, IL, United States},
year={2019}
}

@article{cu26,
title={{California Institute of Technology and United States Geological Survey, Pasadena, Southern California Seismic Network}},
journal={International Federation of Digital Seismograph Networks, Dataset/Seismic Network},
year={1926}
}

@inproceedings{fl95,
author={Faloutsos, Christos and Lin, King-Ip},
title={{FastMap}: A fast algorithm for indexing, data-mining and visualization of traditional and multimedia datasets},
booktitle={Proceedings of the ACM SIGMOD International Conference on Management of Data},
year={1995}
}

@article{gr06,
author={Gibbons, Steven J and Ringdal, Frode},
title={The detection of low magnitude seismic events using array-based waveform correlation},
journal={Geophysical Journal International},
year={2006},
}

@article{mszb19,
author={Mousavi, S Mostafa and Sheng, Yixiao and Zhu, Weiqiang and Beroza, Gregory C},
title={{STanford} {EArthquake} dataset ({STEAD}): A global data set of seismic signals for {AI}},
journal={IEEE Access},
year={2019}
}

@article{mzsb19,
author={Mousavi, S Mostafa and Zhu, Weiqiang and Sheng, Yixiao and Beroza, Gregory C},
title={{CRED}: A deep residual network of convolutional and recurrent units for earthquake signal detection},
journal={Scientific Reports},
year={2019}
}

@article{mezcb20,
author={Mousavi, S Mostafa and Ellsworth, William L and Zhu, Weiqiang and Chuang, Lindsay Y and Beroza, Gregory C},
title={Earthquake transformer—an attentive deep-learning model for simultaneous earthquake detection and phase picking},
journal={Nature Communications},
year={2020}
}

@article{sbi07,
author={Shelly, David R and Beroza, Gregory C and Ide, Satoshi},
title={Non-volcanic tremor and low-frequency earthquake swarms},
journal={Nature},
year={2007}
}

@article{seh16,
author={Shelly, David R and Ellsworth, William L and Hill, David P},
title={Fluid-faulting evolution in high definition: Connecting fault structure and frequency-magnitude variations during the 2014 {Long Valley Caldera}, {California}, earthquake swarm},
journal={Journal of Geophysical Research: Solid Earth},
year={2016}
}

@article{c13,
title={{CalTech} earthquake dataset},
journal={Southern California Earthquake Center},
year={2013}
}

\section*{Acknowledgments}

\par
This work at the University of Southern California is supported by DARPA under grant number HR001120C0157 and by NSF under grant number 2112533. The views, opinions, and/or findings expressed are those of the author(s) and should not be interpreted as representing the official views or policies of the sponsoring organizations, agencies, or the U.S.~Government.

\par
The authors declare that they have no competing interests.

\par
MW and TKSK conceived the general concept of combining FastMap with SVMs and kernel methods for classification of complex objects, independent of~\cite{bka09}. MW also refined the concept in the Earthquake Science domain, implemented the FastMapSVM method presented here, conducted the experiments, and drafted the manuscript. KS and AL conducted various experiments using FastMapSVM in support of those presented here. NN and TKSK provided critical guidance and oversight to the project. AL, TKSK, NN, and KS contributed significantly to manuscript revision.

\par
STEAD data are publicly available at~\url{https://github.com/smousavi05/STEAD}. Ridgecrest data are publicly available at~\url{https://scedc.caltech.edu}.

\clearpage

\section*{Supplementary Material}

Figures S1, S2, S3, and S4.

\clearpage

\renewcommand{\thefigure}{S\arabic{figure}}
\setcounter{figure}{0}

\begin{figure}[!t]
\centering
\includegraphics[width=\textwidth]{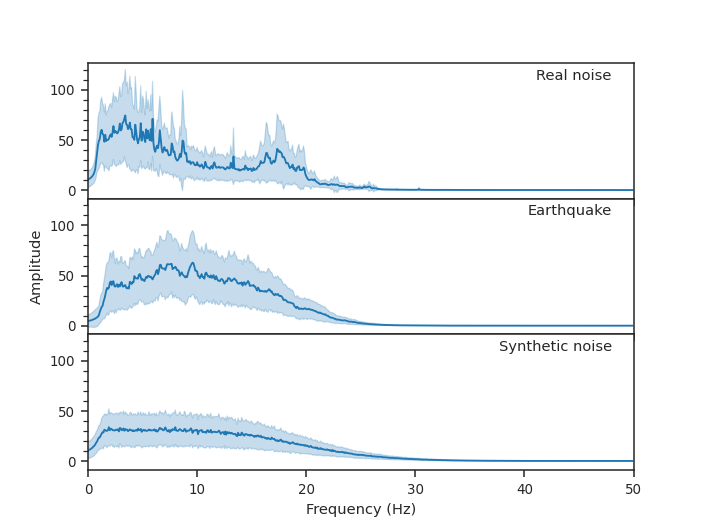}
\caption{\textbf{Average seismogram frequency spectra.} Shows the typical frequency spectra of real noise, earthquake signals, and added synthetic noise.}
\label{fig:suppl_frequency_spectra}
\end{figure}

\begin{figure}[!t]
\centering
\includegraphics[width=\textwidth]{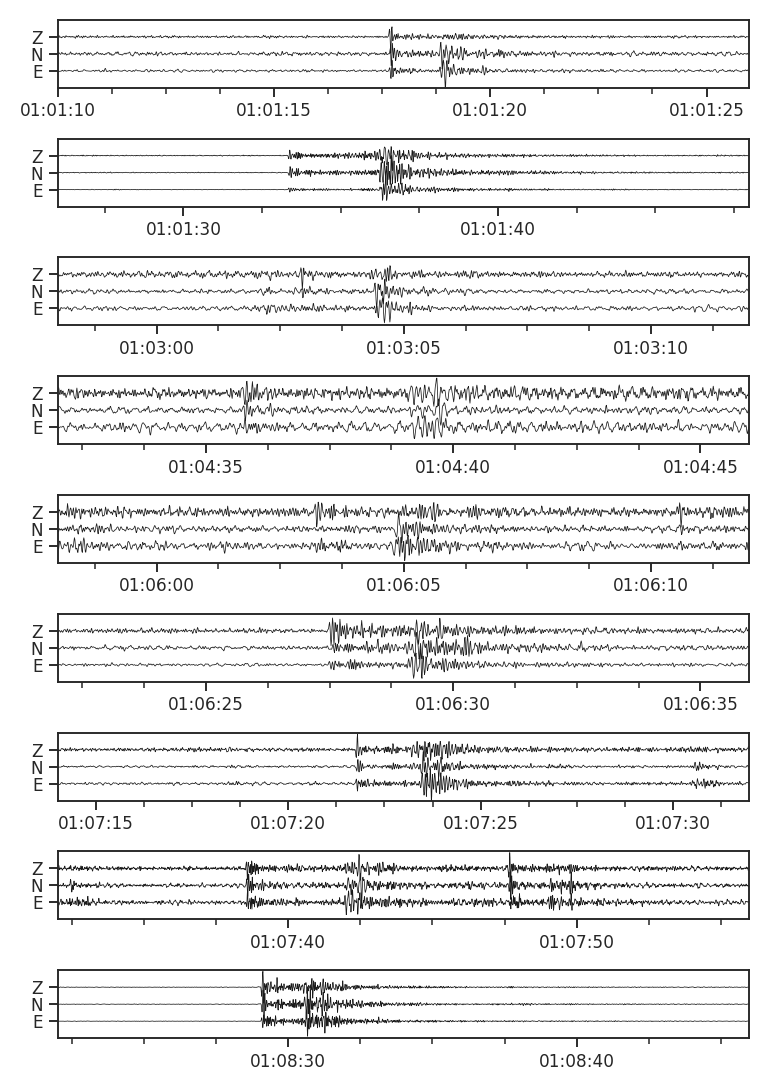}
\caption{\textbf{Earthquakes identified by an automatic FastMapSVM scan on CI.CLC data.} Shows easily discernible earthquake signals.}
\label{fig:suppl_clear}
\end{figure}

\begin{figure}[!t]
\centering
\includegraphics[width=\textwidth]{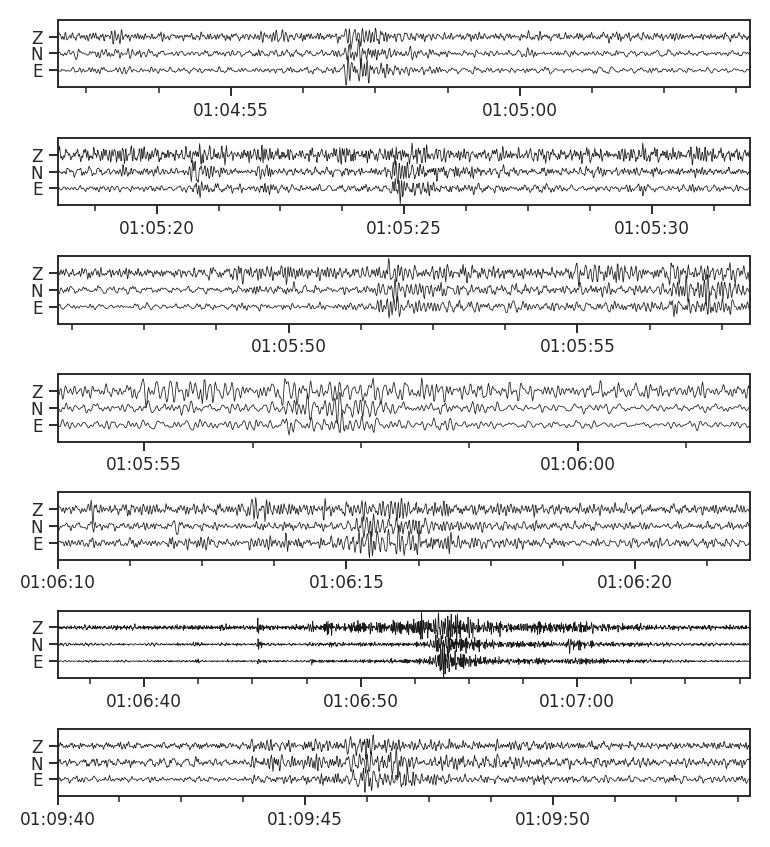}
\caption{\textbf{Potential earthquakes identified by an automatic FastMapSVM scan on CI.CLC data.} Shows earthquake signals with low signal-to-noise ratio.}
\label{fig:suppl_low_snr}
\end{figure}

\begin{figure}[!t]
\centering
\includegraphics[width=\textwidth]{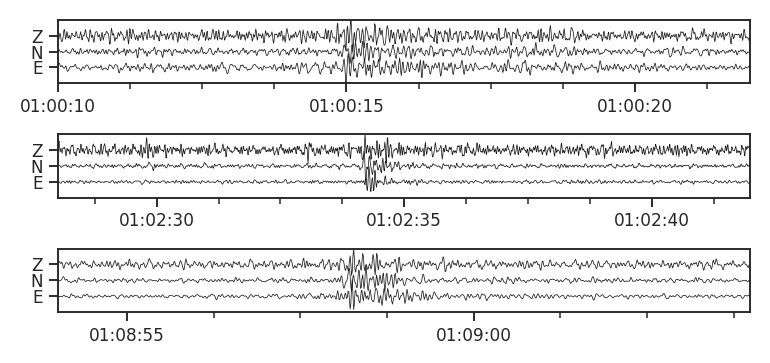}
\caption{\textbf{Ambiguous signals identified as earthquakes by an automatic FastMapSVM scan on CI.CLC data.} Shows ambiguous signals that may or may not be from an earthquake source.}
\label{fig:suppl_ambiguous}
\end{figure}

\end{document}